# Flying Robotics Art: ROS-based Drone Draws the Record-Breaking Mural

Andrei A. Korigodskii[1,2*], Oleg D. Kalachev[3], Artem E. Vasiunik[2,4],
Matvei V. Urvantsev[2], Georgii E. Bondar[3]

*Abstract*— This paper presents the innovative design and successful deployment of a pioneering autonomous unmanned aerial system developed for executing the world's largest mural painted by a drone. Addressing the dual challenges of maintaining artistic precision and operational reliability under adverse outdoor conditions such as wind and direct sunlight, our work introduces a robust system capable of navigating and painting outdoors with unprecedented accuracy. Key to our approach is a novel navigation system that combines an infra-red (IR) motion capture camera and LiDAR technology, enabling precise location tracking tailored specifically for large-scale artistic applications. We employ a unique control architecture that uses different regulation in tangential and normal directions relative to the planned path, enabling precise trajectory tracking and stable line rendering. We also present algorithms for trajectory planning and path optimization, allowing for complex curve drawing and area filling. The system includes a custom-designed paint spraying mechanism, specifically engineered to function effectively amidst the turbulent airflow generated by the drone's propellers, which also protects the drone's critical components from paint-related damage, ensuring longevity and consistent performance. Experimental results demonstrate the system's robustness and precision in varied conditions, showcasing its potential for autonomous large-scale art creation and expanding the functional applications of robotics in creative fields.

## I. Introduction

In recent years, the fusion of robotics with creative arts has opened new possibilities for using drones beyond their traditional roles [1], [2], [3]. Now, these autonomous unmanned aerial systems (UAS) are not just for surveillance or mapping; they're venturing into the world of art, for example, creating wonderful drone shows [4], [5], [6], [7]. This innovative application challenges us to think differently about what drones can do, often demanding higher precision and flexibility in both navigation and control than ever before.

Advances in technology, such as novel positioning systems [8], [9], [10], [11], miniaturization of LiDARs (light detection and ranging devices), and computer vision for spatial recognition, provide the building blocks for this ambitious endeavor. However, applying these technologies to create art with drones introduces unique challenges.

Drones were already used to paint [12], including to paint with spray paint [13], [14], [15], [16]. Moreover, drones were already used to stipple [17], draw graffiti [18], [19] and to create murals [20]. However, these projects mainly took place indoors. The previous record for largest mural drawn by drone was 168 m², and the drawing taken place in partially enclosed hangar [21], where the effects of the weather were limited. In that project, an ultra-wideband radio system was used for positioning.

Our study focuses on a specially designed drone capable of painting the world's largest mural— exceeding 300 m²— autonomously, outdoors and facing different weather conditions, including wind, rain and direct sunlight. We've developed a navigation system that combines infrared (IR) LED lights, an IR camera, and LiDAR, making the drone capable of precise movements even in difficult outdoor conditions. Alongside, we introduce a control system tailored to maintain accuracy essential for drawing, and a custom-designed paint spraying setup that works around the airflow from the drone's propellers.

By exploring these technological developments within our project, this paper aims to show how advanced robotics can be creatively applied to bridge technology and art. The goal is to highlight the potential of drones in new, unconventional roles, encouraging further innovation in the field.

## II. System Design

### A. System architecture

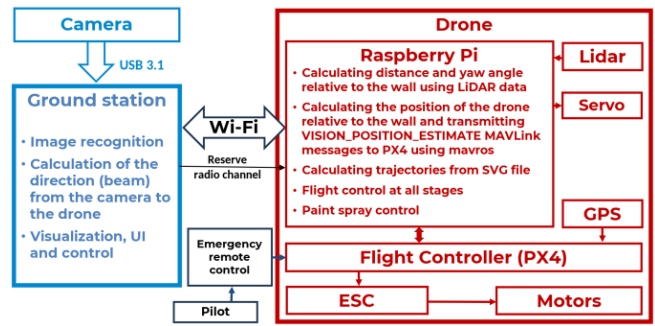

Figure 1. System architecture overview.

The system mainly consists of the unmanned aerial vehicle (UAV), equipped with a single-board companion computer, and a ground station with a tracking camera. An overview of the system is presented in Figure 1.

Robot Operating System (ROS) was used for communication between the parts of the system.

[1] Lomonosov Moscow State University
[2] Sverk Ltd., Moscow
[*] Corresponding author, e-mail: akorigod@gmail.com
[3] Copter Express Technologies Ltd., Moscow
[4] Cognitive Pilot Ltd., Moscow

## B. Localization system

### 1) Localization system overview

To ensure high quality drawing, a high precision localization system was necessary. The volume of the flight space exceeded 1000 m$^3$, necessitating a mean squared error of no greater than 1–2 cm for the localization system to produce aesthetically pleasing images. After reviewing multiple systems (refer to Table I), motion capture (mocap) systems such as VICON or OptiTrack were identified as capable of achieving the necessary precision. However, the financial implications of covering the requisite volume and the costs associated with installing cameras at significant heights were deemed prohibitive.

Nonetheless, the drone's consistent proximity to a large, relatively flat wall presented an opportunity to simplify the localization approach. Utilizing just a single camera, two of the three coordinates (height and lateral movement) could be determined, with the third coordinate, perpendicular to the wall, calculated using an onboard laser rangefinder aimed at the wall.

The flight controller's flight stack accurately calculates the pitch and roll angles, but without external measurement, the yaw angle becomes unreliable due to rapid drift when relying solely on IMU data. In the context of a flat wall, the yaw angle can be effectively determined by analyzing the discrepancy in readings from a pair of onboard laser rangefinders. However, the selected wall for this project featured bricks with significant and deep gaps between them.

To secure precise and reliable measurements of both the distance to the wall and the yaw angle, a 2D LiDAR (Light Detection and Ranging device) was installed onboard. This LiDAR generated a point cloud of the wall, which, through the application of a simple RANSAC algorithm, facilitated accurate calculations of both the distance and the yaw angle.

TABLE I.  COMPARISON OF LOCALIZATION TECHNIQUES

| Technique | Advantages and disadvantages |
|---|---|
| IMU + Baro | + Does not require additional components<br>– Very high drift makes it unusable |
| GNSS RTK or UWB radio | + Low cost<br>– Quite low precision<br>– Susceptible to multipath errors close to the wall |
| Stationary fiducal markers (onboard recognition) | + Low cost<br>± Intermediate precision<br>– Requires heavy onboard processing<br>– Requires placement of multiple fiducial markers |
| Motion capture | + High precision<br>– Extremely high cost to cover necessary volume |
| Presented system: single mocap camera + onboard lidar | + High precision<br>+ Low cost |

### 2) 2D motion capture system

A stationary, high-resolution, low-latency camera with an IR-pass filter was employed to detect IR light-emitting diodes (LEDs) mounted on the UAV. Three LEDs were positioned horizontally on the drone's rear side, creating a pattern of three evenly spaced bright dots recognizable in the camera feed for robust detection, even under direct sunlight. A region-of-interest technique was utilized to enhance recognition speed and reliability further.

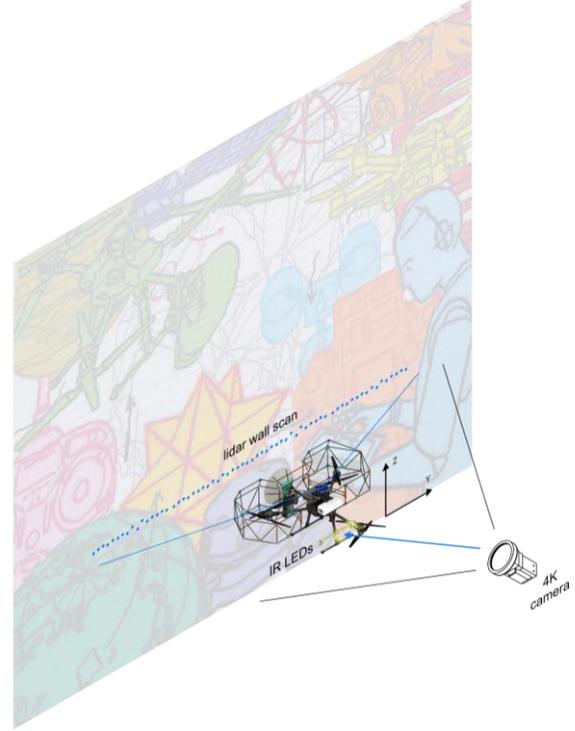

Figure 2.  The positioning system combines information from stationary camera and onboard LiDAR.

### 3) LiDAR

The rotating 2D LiDAR was used to determine the distance and the yaw angle to the wall during the flight. The LiDAR was mounted horizontally (rotation axis was vertical when the drone is standing still) on top of the UAV. The obtained point cloud was approximated by the straight line representing the wall using the RANSAC algorithm.

## C. Trajectory tracking flight control

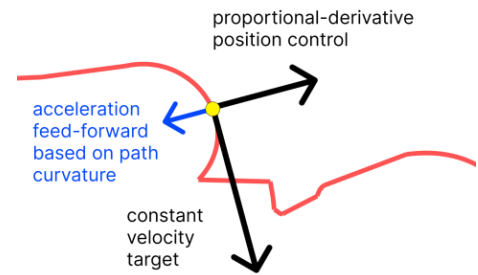

Figure 3.  Trajectory tracking control.

Given the precision required for drawing tasks, control methodologies differed between tangential and normal trajectory tracking. During line drawing, precise position control along the trajectory was not critical; however, maintaining uniform line thickness required the absence of speed oscillations. Therefore, velocity control with a constant target speed was employed during drawing.

Conversely, precise execution of the position setpoint normal to the trajectory was crucial to maintaining drawing accuracy. A proportional–derivative (PD) regulator addressed this need, enabling precise trajectory tracking. The system's nature rendered the integral term unnecessary, with the regulator's output directly feeding into the velocity controller.

Another PD regulator was used to maintain the optimal distance to the wall.

### D. Trajectory planning

To achieve the goal of drawing SVG images provided by the artist, some pre-processing steps were required to make flight trajectories kinematically feasible. Besides outlined below processes for different drawing modes, all lines are extended forwards and backwards by a certain distance (around 30cm). That is done so that the drone has some space to achieve constant target velocity by the point where an actual drawing path starts. In the same way, ending extension is used to slowly decelerate as to not introduce unnecessary oscillations by sharp breaking.

Path extensions were also utilized to achieve precise turning the spray paint on/off in movement. As it takes some time for the servo to engage spray cap and for spray to reach the wall, the control signal to the servo must be given in advance. As such, during movement on extensions, the control program considers the time required to reach "spray start" point based on current position and velocity.

#### 1) Curves drawing

After the SVG image is processed to paths, following post-processing steps are applied.

All lines and curves in paths are iterated upon and either joined or disjoined depending on the tangent angle between them. This is done so that figures that form sharp corners (such as squares) separated into several flight paths, as even with perfect flight control drones cannot make sharp turns while maintaining constant speed. At the same time, figures like sequences of Bézier curves are joined to preserve smoothness during flight and shave off flight time.

Then, paths that are too short (in the order of less than 3–5 cm) and were not joined with others are removed entirely. Finally, resulting paths are sorted to reduce empty travels between paths.

#### 2) Area filling

To fill an area, additional prost-processing of the contours was required in order to generate a set of trajectories forming an infill. Infill paths are generated via following algorithm: a bounding box for the contour area is calculated; then a set of horizontal lines at fixed intervals are interposed on the bounding box, with alternating directions; along the lines, intersections with contours are calculated and used to toggle cropping of the infill lines; to prevent edge cases intersections that are closer than set threshold are ignored; resulting lines are also sorted and reversed during sorting if necessary in case of complex infills with big inner empty areas.

#### 3) Path sorting

Path sorting algorithm maintains a list of all starting and ending points of path segments. The initial point is selected as closest to the bottom center of contours bounding box.

During sorting, remaining points are ranked and the point with minimal rank is chosen to be the next point. If the chosen point was an ending point of a segment, it is reversed.

Ranking formula is:
$$\text{rank} = \text{Dist}_{prev} - z_{start} - z_{end},$$
where:
$\text{Dist}_{prev}$ is distance of a point to the previous point
$z_{start\ t}$ is height of the segment's starting point
$z_{end}$ is is height of the segment's ending point

This method ensures that while travel distances are minimized, the drone will prioritize drawing from bottom to top.

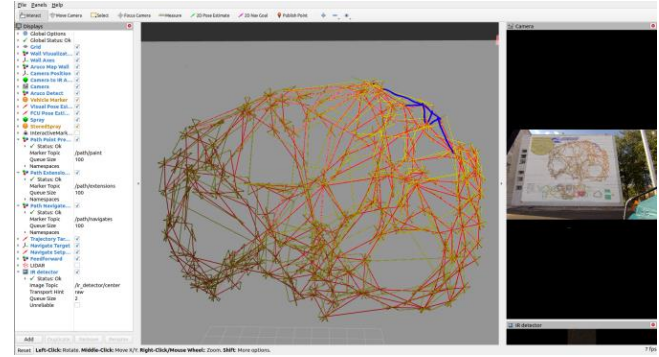

Figure 4. Complex large-scale contours drawing in progress. Saturation of lines indicates order of line drawing. Yellow lines indicate travel without drawing. Short orange lines indicate line extensions. This large figure was intentionally drawn from upper right corner to lower left one by modifying the ranking algorithm.

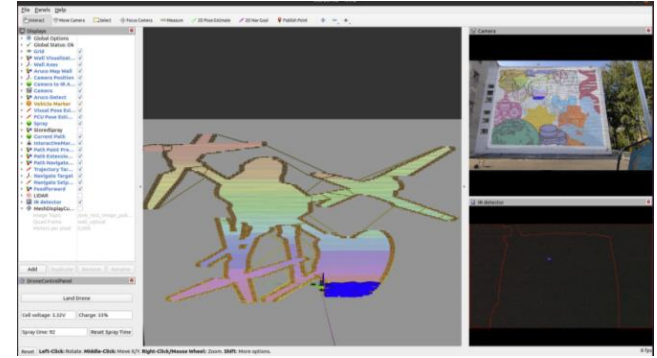

Figure 5. Complex infill drawing in progress: visualization and control GUI in RViz. Hue of infill color indicates order of line drawing. Yellow arrows indicate travel without drawing.

### E. Higher level flight and drawing control

High-level drawing program connected in itself path planning, trajectory tracking, spray paint engagement/disengagement algorithm, emergency, fallback and automation components.

In order to achieve maximal autonomy and safety during operation, high-level flight and drawing control program included following features:

- Voice (TTS and pre-recorded) feedback about actions of the drone; a set of warning about state of

battery, electric current, spray paint runout; positioning system and network lags
- Automatic program termination if flight was interrupted with RC by backup drone operator.
- Autonomous landing procedure to the starting position; triggered either manually via RViz GUI or automatically on detected critically low battery or projected spray paint runout.
- Takeoff and flight to drawing position, as well as autonomous drawing itself— lines to draw can be either selected interactively in RViz or specified as range of path indexes.
- Saving the last painted line and position to enable automatic resuming of drawing from where it was left off.
- Automatic retries of lines where drone could not properly turn on the spray due to high delta between its position and planned trajectory.

High level of automation allowed for drawing flights with zero operator flight interruption or input beyond initial upload of SVG image to draw and selecting drawing paths and parameters.

### F. Connectivity

The ground station and the drone companion computer were connected by Wi-Fi. The Wi-Fi router was located near the wall to maximize the signal strength and connected to the ground station using Ethernet cable.

To ensure robust link to transmit the visual navigation data in case of Wi-Fi failure or delay, a unidirectional backup link was developed. The backup link consisted of a pair of ESP radio modules, connected using ESP-NOW protocol. The ground module was connected to the ground station and transmitted the visual navigation data, which were received by the onboard module, connected to the companion computer. The data was additionally signed to ensure absence of transmission errors or tampering.

### III. SYSTEM INTEGRATION

This section presents how the developed system and autonomous flight algorithms were physically integrated in the drone system as shown in Figure 6.

### A. System requirements

During the design process, the following mechanical requirements were formulated:
1. The drone must carry a 500 g paint canister and navigation equipment.
2. The spray paint nozzle must be close enough to the wall to apply the paint (no more than 5–10 cm).
3. The wall must not prevent the propellers from pulling in enough air to fly.
4. The drone should handle well and have a margin of thrust to withstand the wind.
5. The drone must survive possible contact or collision with the wall.

### B. Mechanical design

The selected drone platform is configured as a coaxial hexacopter. These specific motors arrangement was opted for due to its ability to provide a substantial angular separation between the arms—120 degrees. Such a configuration facilitates the positioning of the spray paint can as close to the drone's center as possible while ensuring sufficient distance from intense airflow. Moreover, the coaxial design contributes to minimizing the interaction area with air, thereby enhancing the drone's stability.

TABLE II. UAV COMPONENT LIST

| Type | Description |
|---|---|
| FCU | CubePilot Cube Orange+ |
| Companion computer | Raspberry Pi 4 Model B, 8 Gb |
| ESCs | X-Rotor 40 A (6 pcs) |
| Motors | T-motor U3 (6 pcs) |
| Propellers | T-motor CF P12x4" (6 pcs) |
| LiDAR | YDLidar X4 |
| IR LEDs | 5W IR LED (3 pcs) |
| Servo | MG946R |
| Frame kit | TAROT 690S |
| Battery | Tattu 6S 10000 mAh LiPo |

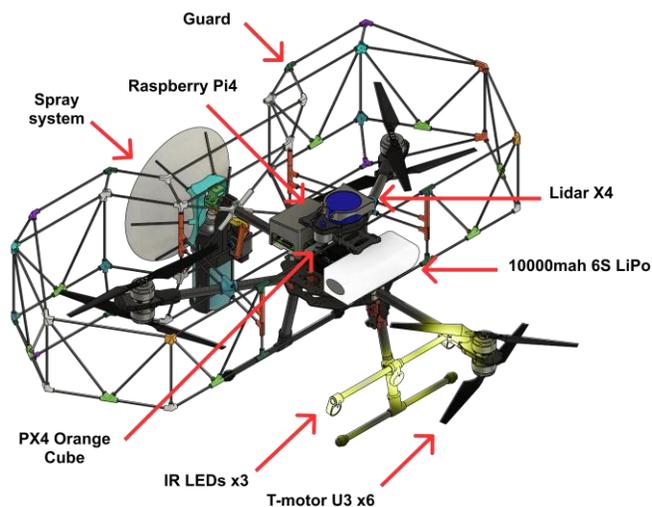

Figure 6. The drone components.

The drone frame consists of a heavily modified Tarot 690S hexacopter kit, a guard, made of 3 mm diameter carbon fiber tubing, and 3D-printed parts.

The high force paint spraying system presses on the can cap with a delay of 100–200 milliseconds. A wide cap with a flat vertical spraying pattern (Montana Flat Jet Wide) was used for filling objects, and a thin one was used for drawing lines.

To protect the paint flow from air currents caused by the propellers and weather conditions, we installed a cone made of cardboard on a carbon tube frame as shown in Figure 7.

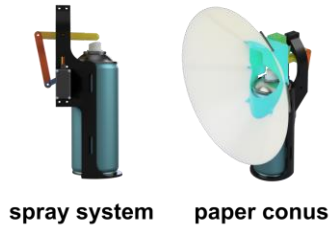

Figure 7. The paint spraying system.

## IV. SYSTEM OPERATIONS

This section describes how the system was operated during the creation of the world's largest drone-drawn mural.

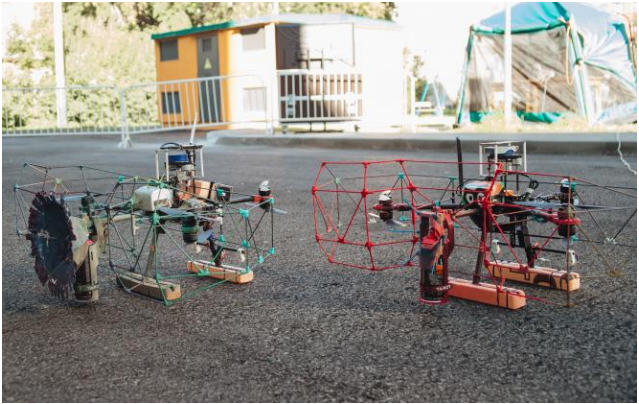

Figure 8. The drones during operations, including a spare one.

### A. System operations basics

The system was operated from the ground control center, located in the tent near the wall where the mural was being created. The camera which tracked the drone was installed on a rigid metal structure near the control center.

The system was operated in the following way:

- SVG file with necessary parts of the image was uploaded to the drone companion computer.
- Drawing mode and settings were configured via launch files or RQT.
- Desired SVG paths were chosen using the interactive graphical interface based on RViz.
- Drone flight script was launched, and the flight sequence and trajectories were confirmed.
- The drone performed flight and drew the selected lines until all the selected lines were drawn, or the battery or the spray can were exhausted, or the landing button was triggered.

In case of drone trajectory tracking failures or operator errors leading to errors in the drawing operators used the drone in "eraser mode". As RViz visualization contained real spraying history as well as planned trajectories visualization reprojected onto the camera image, operators were able to pinpoint erroneous drawing segments and place "erasing" segments over them, which were then drawn in corresponding foreground color.

### B. Recognition system calibration

To calculate the camera location with respect to the wall, and to ensure the repeatability of the drawing throughout the long project duration regardless of possible camera shifts, in the beginning of each day the calibration procedure was performed. Four large ArUco markers were placed on the wall, and their exact positions were measured. During the calibration procedure, a single picture was taken with the camera with IR-pass filter taken off. By locating ArUco markers on the image with their coordinates known, the camera's position and orientation was calculated.

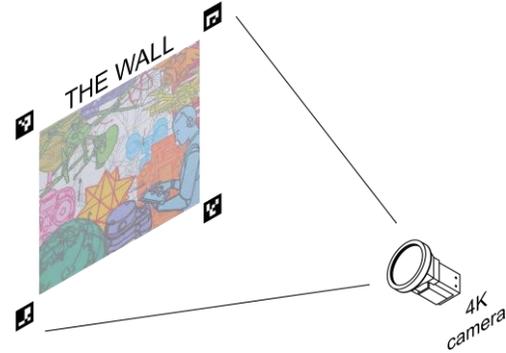

Figure 9. The calibration procedure.

### C. System maintenance

After each flight, the exhausted spray can and/or an exhausted battery were replaced by the ground engineering team. If necessary, drone propellers were cleaned from the paint deposited on them.

## V. CONCLUSION

This study demonstrates the successful design and deployment of an autonomous unmanned aerial system for creating the world's largest outdoor mural drawn by a drone. Our work addresses several key challenges in applying drone technology to large-scale, high-precision artistic endeavors in uncontrolled outdoor environments. The developed system integrates a novel navigation approach combining infrared (IR) LEDs, an IR camera, and a 2D LiDAR to achieve precise localization in the presence of environmental disturbances such as wind, rain, and varying lighting conditions. Furthermore, we developed an innovative control system architecture that applies different controls for movement along and perpendicular to the drawing trajectory, ensuring the precision necessary for artistic rendition.

While the current implementation is highly effective, several avenues for future research and development remain. Enhancing the system's adaptability to extreme weather conditions and optimizing painting efficiency could improve system's performance and simplify its use. We plan to investigate the possibilities of powering the drone by wire, which could provide a continuous power supply and extend operation times. We also aim to explore the option to deliver paint to a drone via a tube from the ground instead of using aerosol paints, allowing for a more consistent supply, as well as investigate the options that can utilize multiple colors simultaneously. Finally, the application of machine learning techniques for dynamic path planning could increase the system's efficiency.

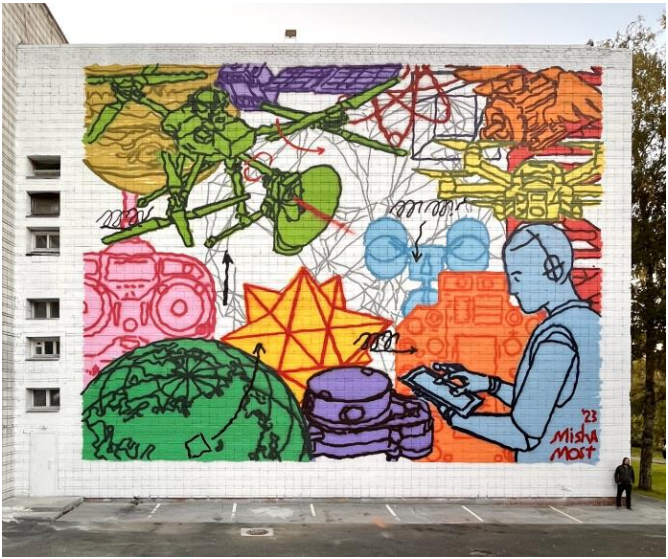

Figure 10. The resulting mural.

APPENDIX

The PDF document containing additional images of the system and its user interface is available at: https://drive.google.com/file/d/1X0xhdx9ANM1YkDodqMsnP_bYnjGLvQUw/view.

Additional information about the project as well as extra images are also available at https://artomatika.com/en/mural.

ACKNOWLEDGMENT

The authors would like to thank Universal University and 20.35 University for their financial and organizational support, which was instrumental for this study.

The authors thank the artistic director of the project, Misha Most, for his invaluable help and artistic guidance throughout the project.